\documentclass[10pt,leqno]{amsart}
\usepackage{graphicx}
\usepackage{float}       
\usepackage{tabularx}    
\usepackage{booktabs}
\usepackage{array}
\usepackage{multirow}
\usepackage{indentfirst}
\usepackage{csquotes}
\usepackage{amssymb}
\usepackage{amsthm}
\usepackage{amssymb}

\usepackage{amsmath}
\usepackage[numbers,sort&compress]{natbib}
\setcitestyle{numbers,square,sort&compress}

\newcolumntype{L}{>{\raggedright\arraybackslash}X}
\newcolumntype{C}{>{\centering\arraybackslash}X}
\newcolumntype{R}{>{\raggedleft\arraybackslash}X}

\usepackage{xcolor}
\usepackage{paralist}
\usepackage{hyperref}
\usepackage{fancyhdr}
\usepackage{etoolbox}

\usepackage{graphicx}
\graphicspath{{figures/}}

\AtBeginDocument{%
    \setlength{\abovedisplayskip}{4pt plus 1pt minus 1pt}%
    \setlength{\belowdisplayskip}{4pt plus 1pt minus 1pt}%
    \setlength{\abovedisplayshortskip}{2pt}%
    \setlength{\belowdisplayshortskip}{3pt plus 1pt minus 1pt}%
}

\hypersetup{
    colorlinks=true,
    linkcolor=black,
    citecolor=black,
    filecolor=black,
    urlcolor=black
}

\usepackage{fancyhdr}

\title{Decomposed Relation Extraction from Aviation Accident Narratives: Evaluating Transformer-Based Safety Knowledge Discovery under Relational Sparsity}

\author{
Aziida Nanyonga$^{1,*}$,
Mustafa Tahsin Yilmaz$^{2}$,
Saad Ezzini$^{3}$,
Ugur Turhan$^{4}$,
and Graham Wild$^{4}$
}

\date{}

\begin{document}

\begin{abstract}
Relation extraction from aviation accident narratives is challenging because genuine safety relations are sparse within the large space of possible concept pairs. This study investigates this problem using a decomposed framework that separates relation classification from concept identification. A 300-narrative National Transportation Safety Board (NTSB) benchmark containing 2,524 unique reference concept-text nodes was exhaustively expanded into 21,038 ordered candidate pairs, of which only 248 (1.18\%) represented \texttt{RESULTS\_IN} or \texttt{CONTRIBUTES\_TO} relations. A term frequency--inverse document frequency (TF--IDF) Linear Support Vector Machine (SVM) and a weighted RoBERTa classifier were developed on an independent graph-annotated resource using hard-negative sampling and frozen before evaluation on the sealed benchmark. Development Macro-F1 scores of 0.4738 for the SVM and 0.4809 for RoBERTa decreased to 0.2741 and 0.2420, respectively, under exhaustive evaluation, with corresponding Matthews correlation coefficients (MCCs) of 0.0128 and 0.0197. False-positive proliferation was the dominant failure mode: the SVM recovered 96 reference relations while generating 7,018 false positives, whereas RoBERTa recovered 125 while generating 9,824. Post-test diagnostics revealed substantial overlap between RoBERTa confidence distributions for correct and incorrect predictions and extensive false positives within category directions that also contained genuine relations. A complementary end-to-end SmolLM2 evaluation further revealed missed relations, unsupported relations, and malformed structured outputs. These findings show that correct concept endpoints alone are insufficient for precise relation extraction and that performance under sampled-negative development conditions may not characterize behavior in an exhaustive, naturally sparse candidate space. Candidate-space management and false-positive control therefore remain central challenges for structured knowledge discovery from aviation safety narratives.
\end{abstract}

\maketitle

\begin{center}
\small
$^{1}$SDAIA-KFUPM Joint Research Center for Artificial Intelligence,
King Fahd University of Petroleum and Minerals (KFUPM),
Dhahran 31261, Saudi Arabia

\medskip
$^{2}$Department of Industrial Engineering, Faculty of Engineering, King Abdulaziz University, Jeddah 21589, Saudi Arabia

\medskip
$^{3}$Department of Information and Computer Science, King Fahd University of Petroleum and Minerals, Dhahran 31261, Saudi Arabia

\medskip
$^{4}$Capability Systems Centre,
University of New South Wales,
Canberra, ACT 2600, Australia

\medskip
$^{4}$Capability Systems Centre,
University of New South Wales,
Canberra, ACT 2600, Australia

\medskip
$^{*}$Corresponding author:
\href{mailto:aziida.nanyonga@kfupm.edu.sa}{aziida.nanyonga@kfupm.edu.sa}
\end{center}

\medskip

\noindent\textbf{Keywords:} aviation safety; relation extraction; natural language processing; text mining; transformer models; knowledge discovery; relational sparsity; NTSB accident narratives

\bigskip

\section{Introduction}
\label{sec:introduction}

Aviation accident investigation reports contain detailed textual accounts of events, operational conditions, technical failures, human actions, environmental influences, and their interactions. Although structured database fields support conventional statistical analysis, narrative sections preserve contextual and relational information that cannot always be represented adequately through predefined variables alone. Recent natural language processing (NLP) research has consequently investigated aviation narratives using text mining, Transformer-based information extraction, large language models (LLMs), retrieval-augmented generation (RAG), and knowledge graphs to transform unstructured reports into representations that can support large-scale safety analysis \cite{Orviz2026NLP,Ren2025RACI}.

An important objective within this research direction is the extraction of relationships between safety-relevant concepts. Identifying that a narrative contains a technical failure, environmental condition, operational event, or consequence provides useful information, but structured safety analysis also requires understanding how those concepts are connected. Relations can encode, for example, whether one condition contributes to another event or whether an event results in a subsequent outcome. Such relational representations provide the basis for constructing structured safety knowledge, including knowledge graphs in which concepts are represented as nodes and their relationships as directed edges. Aviation knowledge graphs have already been explored for information retrieval, question answering, causal analysis, and organization of accident knowledge \cite{Xu2024CausationKG,Chen2024AviationKG}.

Recent studies demonstrate substantial progress toward extracting causal and relational information from aviation reports. Transformer-based methods have been developed to identify causal spans and structures from NTSB narratives, while LLM-based approaches increasingly generate causal factors, relations, and knowledge-graph structures directly from accident or incident reports \cite{Chen2026ARCEM,Basil2025LLM}. Other work has combined prompt engineering and retrieval-augmented generation to identify causation factors and their interrelationships for aviation accident knowledge-graph construction \cite{Xiang2026RAGKG}. These developments indicate growing interest in moving beyond document classification or topic identification toward explicit representations of relationships contained in safety narratives.

However, relation extraction introduces a candidate-space problem that is distinct from concept identification. Given $n$ identified concepts in a narrative, an unconstrained directed relation classifier may need to consider

\begin{equation}
N_{\mathrm{pairs}} = n(n-1)
\end{equation}

ordered source--target pairs. Only a small proportion of these combinations may correspond to genuine relations. This produces a highly imbalanced prediction space in which non-relational pairs can greatly outnumber positive relations. Severe negative-class dominance is also recognized more generally in relation-extraction research, where large candidate spaces can contain overwhelmingly more unrelated entity pairs than positive relational instances \cite{Safavi2022GraphLearning}. Under these conditions, even a comparatively small tendency to assign positive labels to negative candidates can produce a large absolute number of unsupported relations.

The sparsity of genuine relations within the candidate space has important implications for model development. Negative examples can be used to provide contrastive supervision, while hard-negative mining can further expose models to semantically similar but incorrect relation candidates that are more difficult to distinguish from valid relations \cite{Olasunkanmi2025RELATE}. Nevertheless, performance obtained under a sampled candidate distribution does not necessarily characterize model behavior when every possible pair is considered. For knowledge-graph construction, the latter setting is especially relevant because false-positive classifications become unsupported graph edges and can accumulate rapidly as the candidate space expands.

A further challenge is determining where errors originate within an end-to-end extraction pipeline. If a model fails to recover a relation directly from narrative text, the error may result from failure to identify the relevant concepts, incorrect pairing of otherwise valid concepts, incorrect relation classification, incorrect directionality, or failure to generate a valid structured representation. Evaluating only the final extracted structure makes these sources of error difficult to separate. A decomposed evaluation in which reference concepts are supplied as reference endpoints provides a controlled means of isolating relation discrimination from concept discovery.

Motivated by these considerations, this study investigates directed relation extraction from National Transportation Safety Board (NTSB) aviation accident narratives under relational sparsity. A manually annotated benchmark of 300 narratives is represented using eight safety-concept categories, from which unique concept-text nodes are constructed. Rather than filtering candidate pairs using relation patterns observed in the benchmark, all possible ordered pairs of reference concepts are retained. This produces 21,038 candidate relations, of which only 248 (1.18\%) correspond to reference \texttt{RESULTS\_IN} or \texttt{CONTRIBUTES\_TO} relations.

Two discriminative model families are evaluated within this framework: a TF--IDF Linear Support Vector Machine (SVM) and a RoBERTa-based contextual relation classifier. Model development is performed using an independent graph-annotated resource with safe hard-negative sampling and narrative-level separation between training and validation data. Model configuration and selection are completed before the sealed benchmark is evaluated. This design enables comparison between performance under the controlled development distribution and performance when the frozen classifiers are exposed to the complete naturally sparse candidate space.

The study additionally examines previously generated SmolLM2-1.7B-Instruct outputs on the same 300 narratives under zero-shot, one-shot, few-shot, and retrieval-grounded prompting. This component provides a complementary end-to-end perspective in which the language model must discover relation endpoints and generate structured relations directly from narrative text. Because this task differs from gold-endpoint candidate classification, the generative and discriminative results are analyzed as complementary error profiles rather than as directly comparable model scores.

This study therefore addresses the following research questions:

\begin{enumerate}
    \item[\textbf{RQ1.}] How effectively can classical and Transformer-based classifiers identify directed semantic relations between known safety concepts in aviation accident narratives?

    \item[\textbf{RQ2.}] How does relation-classification performance change when models developed using controlled hard-negative sampling are evaluated over the naturally sparse exhaustive candidate space?

    \item[\textbf{RQ3.}] How do relation-extraction errors vary across relation types, concept-category directions, and model architectures?

    \item[\textbf{RQ4.}] What complementary limitations emerge between end-to-end generative relation extraction and decomposed gold-endpoint relation classification on the same aviation safety benchmark?
\end{enumerate}

The main contributions of this study are fourfold:

\begin{enumerate}
    \item A decomposed, direction-aware framework for aviation safety relation extraction that separates concept endpoints, ordered candidate-pair construction, and contextual relation classification.

    \item A controlled evaluation protocol that separates relation discrimination from concept-identification error by evaluating relation classification using gold concept endpoints.

    \item An exhaustive evaluation under the naturally sparse candidate distribution, revealing a substantial generalization gap between controlled hard-negative development conditions and deployment-like sparse relation extraction.

    \item A complementary analysis of an existing end-to-end generative LLM approach on the same 300 narratives and reference relation set, showing that relation extraction remains challenging under both decomposed and generative formulations while revealing different sources of error. Because the two formulations represent different prediction tasks, their results are interpreted as complementary error profiles rather than as directly comparable model scores.
\end{enumerate}

The remainder of this paper is organized as follows. Section~2 reviews related work on aviation safety NLP, causal and relational information extraction, knowledge-graph construction, and candidate-pair imbalance. Section~3 describes the study design, datasets, relation ontology, candidate-generation procedure, model development, and evaluation framework. Section~4 presents the experimental results, including development and sealed-benchmark performance, error analysis, confidence analysis, and cross-model comparisons. Section~5 discusses the findings in relation to relational sparsity, candidate-space construction, false-positive control, and the complementary end-to-end generative evaluation. Finally, Section~6 summarizes the main conclusions, limitations, and directions for future research.

\section{Related Work}
\label{sec:related_work}

\subsection{Natural Language Processing for Aviation Safety Narratives}

Aviation safety databases contain large collections of narrative reports that complement structured accident and incident attributes with descriptions of operational events, human actions, technical conditions, environmental factors, and investigative findings. The scale and unstructured nature of these collections has motivated increasing use of natural language processing (NLP) for automated safety analysis. Earlier approaches emphasized tasks such as text classification, information retrieval, topic modelling, and text mining \cite{Tanguy2016AviationNLP}, whereas more recent studies have applied machine learning, deep learning, and pretrained language models to tasks such as injury-severity prediction, occurrence classification, flight-phase classification, and causal information extraction from aviation safety narratives \cite{Chen2026ARCEM,Nanyonga2025AviationNLP,Jonk2023AviationNLP,Nanyonga2026FlightPhase}.

Studies using Australian aviation safety narratives have demonstrated the utility of NLP for predicting injury levels, classifying aviation safety incidents, and identifying flight phases \cite{Nanyonga2025AviationNLP,Nanyonga2025DeepLearning,Nanyonga2026FlightPhase}. Other investigations have employed probabilistic and matrix-factorization-based topic-modelling techniques to explore recurring themes in aviation incident narratives and examine how methodological choices influence their interpretation \cite{Nanyonga2025TopicModels,NanyongaTopicAssessment}. Although these approaches provide valuable document-level predictions and thematic representations, they do not directly establish directed semantic relationships between individual safety concepts. This distinction motivates further investigation of relation extraction as a complementary level of aviation narrative analysis.

This progression is particularly relevant to accident investigation because important safety information is often expressed through relationships rather than isolated concepts. A narrative may contain multiple technical, operational, environmental, and human factors, but identifying these factors independently does not establish how they interact within an occurrence. Consequently, aviation NLP has increasingly extended beyond document-level categorization toward causal information extraction, relation extraction, and structured knowledge representation.

\subsection{Causal and Relational Information Extraction from Aviation Reports}

Recent studies have explicitly addressed causal information extraction from aviation narratives. Chen et al. proposed the Aviation Reports Causality Extraction Model (ARCEM), which combines BERT-based contextual representations, multi-granular embeddings, a Transformer architecture, and a conditional random field decoder to extract cause, effect, and embedded-causality spans from NTSB causal narratives \cite{Chen2026ARCEM}. This work demonstrates the feasibility of representing aviation-specific causal information through supervised sequence labeling.

Generative approaches have subsequently extended aviation causal extraction beyond token-level supervised models. Ren et al. introduced a Retrieval-Augmented Generation-aided Causal Identification (RACI) framework that combines retrieval and generation to identify causal information from aviation accident reports \cite{Ren2025RACI}. Retrieval-augmented approaches have also been used to extract causation factors and their interrelationships from NTSB reports and organize them into structured causal representations for subsequent analysis \cite{Xiang2026RAGKG}.

These approaches represent an important transition from identifying individual safety factors toward recovering relationships among them. However, causal-span identification and end-to-end generation combine several stages of the extraction problem, including concept discovery, endpoint selection, relation assignment, and, in some settings, structured-output generation. Consequently, their final outputs do not necessarily isolate the ability of a model to discriminate a directed semantic relation once its endpoints are already known. The present study focuses specifically on this relation-discrimination problem while retaining an end-to-end generative evaluation as a complementary perspective.

\subsection{Large Language Models and Aviation Safety Knowledge Graphs}

Knowledge graphs provide a structured representation of safety information in which concepts are represented as nodes and semantic or causal relationships as edges. Recent aviation research has increasingly combined LLM-based information extraction with knowledge-graph construction to transform narrative reports into machine-readable relational representations.

LLM-based methods have been used to construct aviation accident causation knowledge graphs by extracting accident-related entities and their relationships from report collections \cite{Chen2024AviationKG}. Other approaches have integrated prompt engineering, retrieval-augmented generation, relation normalization, and graph-based analysis to support extraction and reasoning across aviation accident or incident narratives \cite{Xiang2026RAGKG,Cheng2026ASHCKG}. Hybrid KG--LLM frameworks have further combined explicit relational representations with generative reasoning to support interpretable analysis of accident mechanisms \cite{Lin2026AI4AirSafe}.

These developments demonstrate the potential of relational representations for cross-report analysis, retrieval, causal-chain exploration, and safety-oriented reasoning. They also increase the importance of relation quality. When extracted relations are incorporated into a knowledge graph, false-positive predictions become unsupported edges and may affect graph connectivity, retrieval, or downstream reasoning. Relation precision and false-positive behavior are therefore important considerations when NLP outputs are used to construct structured safety knowledge.

\subsection{Candidate-Pair Imbalance and Negative Sampling in Relation Extraction}

A central challenge in relation extraction is the imbalance between the relatively small number of related entity pairs and the much larger number of possible unrelated pairs. As the number of candidate concepts increases, exhaustive pair construction can produce a rapidly expanding negative candidate space. This imbalance can cause model development to be dominated by easy negative instances and can make unrestricted evaluation sensitive to false-positive predictions.

Relation-extraction research has consequently employed negative sampling to construct informative negative examples and improve discrimination between related and unrelated representations \cite{Xu2022NegativeSampling}. Hard-negative mining provides a more targeted strategy by exposing models to difficult negative examples that more closely resemble valid relation candidates, thereby encouraging finer discrimination between relational and non-relational pairs \cite{Olasunkanmi2025RELATE}.

Although these strategies can improve model development, negative sampling also changes the relation prevalence encountered during training and validation. Performance measured under a controlled positive-to-negative ratio may therefore differ from performance when a classifier is applied to the complete candidate space of a document. This distinction is especially important in sparse relation-extraction settings, where even a modest rate of positive predictions among negative candidates can produce a large absolute number of unsupported relations.

\subsection{Research Gap and Positioning of the Present Study}

Existing aviation NLP research has demonstrated the feasibility of extracting causal spans, identifying safety factors, generating causal relationships, and constructing aviation safety knowledge graphs using Transformer models, LLMs, retrieval-augmented generation, and hybrid KG--LLM architectures \cite{Chen2026ARCEM,Ren2025RACI,Chen2024AviationKG,Xiang2026RAGKG,Cheng2026ASHCKG,Lin2026AI4AirSafe}. In parallel, broader representation-learning and relation-extraction research has highlighted the value of negative sampling and hard-negative selection for improving discrimination between positive and negative instances \cite{Xu2022NegativeSampling,Olasunkanmi2025RELATE}.

A comparatively less examined issue is how relation classifiers developed under sampled negative conditions behave when evaluated exhaustively over the naturally sparse candidate space of complete narratives. End-to-end extraction can obscure this issue because its errors may arise from concept identification, endpoint selection, relation classification, directionality, or structured generation. Conversely, sampled candidate-pair evaluations may not reveal the cumulative false-positive behavior that emerges when every ordered concept pair is considered.

The present study addresses this gap through a decomposed, direction-aware evaluation in which gold concept endpoints are supplied to the classifiers and all ordered pairs of distinct concept-text nodes are retained during sealed evaluation. This design isolates relation discrimination from concept-identification error while preserving the naturally sparse candidate distribution. By comparing classifiers developed with controlled hard-negative sampling against exhaustive sealed evaluation, and by examining existing end-to-end generative outputs on the same aviation benchmark as a complementary task formulation, the study investigates candidate-space sparsity and false-positive control as distinct challenges in aviation safety relation extraction.

\section{Materials and Methods}
\label{sec:methods}

\subsection{Study Design and Evaluation Corpus}
\label{sec:study_design}

This study investigated directed semantic relation extraction from aviation accident narratives using a decomposed experimental framework that separated concept identification from relation classification. The primary evaluation supplied reference safety concepts as relation endpoints and required the classifiers to assign each ordered concept pair to one of three classes: \texttt{NO\_RELATION}, \texttt{RESULTS\_IN}, or \texttt{CONTRIBUTES\_TO}. Relation direction was preserved, such that $(c_i,c_j)$ and $(c_j,c_i)$ were treated as distinct candidate instances.

The primary evaluation corpus comprised 300 National Transportation Safety Board (NTSB) accident narratives with manually established safety concepts, supporting textual evidence, and directed semantic relations. The benchmark was reserved for final evaluation and was not used for classifier training, hyperparameter selection, negative-sample construction, threshold optimization, or model selection. Model development instead used an independently annotated aviation safety graph resource, with overlapping narratives removed before classifier development.

Two discriminative approaches were evaluated: a term frequency--inverse document frequency (TF--IDF) representation with a linear support vector machine (SVM) and a contextual classifier based on RoBERTa-base \cite{Liu2019RoBERTa}. Both classifiers were developed independently of the sealed benchmark and frozen before final evaluation. Previously generated SmolLM2-1.7B-Instruct outputs for the same 300 narratives and reference relation set were additionally analyzed as a complementary end-to-end formulation in which relation endpoints and relations were generated directly from narrative text.

\begin{figure}[H]
\centering
\includegraphics[width=\textwidth]{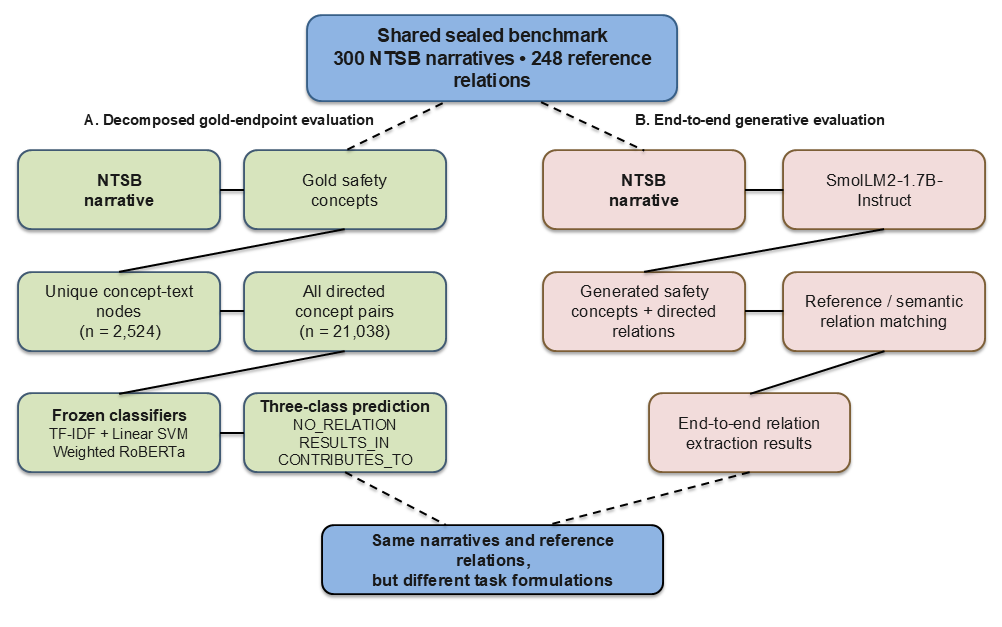}
\caption{Experimental framework for aviation safety relation extraction. The decomposed evaluation uses reference safety concepts to construct exhaustive ordered concept pairs for relation classification, whereas the end-to-end generative evaluation requires SmolLM2-1.7B-Instruct to generate concepts and relations directly from the narrative. Both settings use the same 300-narrative benchmark and reference relation set but represent different task formulations.}
\label{fig:framework}
\end{figure}

Figure~\ref{fig:framework} summarizes the experimental framework. The decomposed pathway isolates candidate-level relation discrimination under known concept endpoints, whereas the generative pathway incorporates concept discovery, relation generation, and structured-output generation.

\unskip

\subsection{Reference Concepts, Relation Ontology, and Candidate Construction}
\label{sec:reference_benchmark}

The 300-narrative benchmark contained 2,990 reference concept annotations representing safety-relevant entities and events. Identical normalized concept strings within the same narrative were consolidated into single concept-text nodes while retaining all associated category memberships as metadata, producing 2,524 unique reference concept-text nodes. Concept-category metadata were retained for descriptive and post-test analyses but were not supplied to the discriminative classifiers as predictive features.
For a narrative containing $n$ unique reference nodes, all ordered pairs of distinct nodes were retained:

\begin{equation}
P=\{(c_i,c_j)\mid c_i,c_j\in C,\;i\neq j\},
\qquad |P|=n(n-1).
\end{equation}

Both directions were retained because the target relations were directional. Each candidate was represented using the source concept text, target concept text, and associated narrative context. Relation classification used three directed labels. \texttt{RESULTS\_IN} represented a relation in which the source concept directly led to, produced, or resulted in the target concept. \texttt{CONTRIBUTES\_TO} represented a contributory relation without requiring a direct result, while \texttt{NO\_RELATION} indicated that neither target relation existed from the source to the target concept.

Reference relations were mapped to the consolidated concept-text nodes in the same direction as their annotations. Exhaustive candidate construction across the 300 narratives produced 21,038 ordered pairs: 228 \texttt{RESULTS\_IN}, 20 \texttt{CONTRIBUTES\_TO}, and 20,790 \texttt{NO\_RELATION} instances. Thus, only 248 candidate pairs (1.18\%) represented positive relations, producing the naturally sparse candidate distribution used for final evaluation.

Retaining the unconstrained candidate space across the entire narrative serves two deliberate methodological purposes. First, causal chains in aviation accident reports frequently transcend sentence boundaries and paragraph breaks (for example, an antecedent maintenance oversight or weather hazard mentioned early in a narrative resulting in an in-flight control excursion paragraphs later), meaning strict intra-sentence or local window filtering risks prematurely discarding valid long-range safety relations. Second, evaluating unconstrained candidate pairs establishes an empirical upper bound on combinatorial candidate explosion and false-positive accumulation, exposing the baseline challenge before any heuristic pruning is applied.

\subsection{Development Dataset and Hard-Negative Sampling}
\label{sec:development_dataset}

Classifier development used an independently annotated aviation safety graph resource containing 140 narratives. Narrative identifiers were compared with those in the 300-narrative evaluation benchmark, and five overlapping narratives were removed, leaving 135 development narratives. The remaining narratives were divided at the narrative level into 108 training and 27 validation narratives, ensuring that candidate pairs from the same narrative could not occur in both partitions.

Relation annotations were harmonized to the target ontology, yielding 287 positive \texttt{RESULTS\_IN} or \texttt{CONTRIBUTES\_TO} instances. Negative candidates were restricted to ordered node pairs for which no annotated relation of any type existed in the source graph, thereby avoiding the treatment of non-target annotated relations as negatives.

Because exhaustive pair construction generated substantially more negative than positive candidates, controlled hard-negative sampling was used during development. Negative sampling provides informative negative instances from larger candidate spaces \cite{Xu2022NegativeSampling}, while hard-negative strategies can emphasize examples that are more difficult to distinguish from genuine relations \cite{Olasunkanmi2025RELATE}. A negative candidate was considered structurally plausible when its source--target concept-category direction was also observed among positive relations in the development data. No category-direction information from the sealed benchmark was used for this procedure.

A 2:1 negative-to-positive ratio was applied, resulting in 574 hard-negative instances for the 287 positive relations and a total development set of 861 candidate pairs. The narrative-level partition contained 687 training and 174 validation instances. Hard-negative sampling was used only during development; no negative subsampling or category-based candidate filtering was applied during final evaluation.

\subsection{Relation Classification Models}
\label{sec:relation_models}

Two discriminative classifiers were developed using the same narrative and ordered source--target concept information. No concept-category metadata, manually annotated evidence spans, or benchmark-derived structural rules were supplied as predictive features.

The first classifier used TF--IDF features with a linear SVM as a classical sparse-text baseline. Each candidate instance combined the complete narrative with explicit source and target concept markers. TF--IDF features consisted of unigrams and bigrams, with lowercase conversion, a minimum document frequency of 2, a maximum document frequency of 0.95, sublinear term-frequency scaling, and a maximum vocabulary size of 100,000 features. The resulting training vocabulary contained 14,888 features. Classification used class-balanced weighting, a fixed random seed of 42, and a regularization parameter of $C=5$, selected on the development validation partition using Macro-F1.

The second classifier used RoBERTa-base \cite{Liu2019RoBERTa}. Candidate inputs were structured as the source concept, target concept, and complete narrative, in that order. Inputs were tokenized to a maximum sequence length of 512 tokens; when truncation was required, the narrative end was truncated so that both relation endpoints remained available to the model. The median narrative length across the benchmark was approximately 210 words, meaning the majority of narratives fitted comfortably within the 512-token context window without truncation. For longer reports exceeding this limit, placing the relation endpoints at the sequence head ensured that the candidate concepts themselves remained fully preserved in the input representation, although context describing distant subsequent events in lengthy narratives was subject to right-side narrative truncation.

RoBERTa was trained using weighted cross-entropy, with class weights computed exclusively from the training partition. The resulting weights were 0.5000 for \texttt{NO\_RELATION}, 1.9244 for \texttt{RESULTS\_IN}, and 2.0818 for \texttt{CONTRIBUTES\_TO}. A development-only sensitivity analysis considered weighting strengths $\alpha \in \{0,0.5,0.75,1.0\}$, where $\alpha=0$ represented unweighted loss and $\alpha=1$ represented the full balanced class weights. Full weighting ($\alpha=1$) was selected using validation Macro-F1.

The final RoBERTa configuration used a learning rate of $2\times10^{-5}$, a mini-batch size of 8, weight decay of 0.01, a warm-up proportion of 0.10, and a fixed random seed of 42. Model checkpoints were evaluated on the development validation partition, and the epoch-5 checkpoint was selected according to validation Macro-F1 and frozen before sealed-benchmark inference.

For both classifiers, all preprocessing, weighting, and model-selection decisions were based exclusively on the development data. No information from the sealed 300-narrative benchmark was used for parameter estimation, preprocessing decisions, model selection, class weighting, or threshold adjustment.
\subsection{Model Selection and Sealed Evaluation}
\label{sec:sealed_evaluation}

Following development, the selected TF--IDF + Linear SVM and RoBERTa processing pipelines were frozen before evaluation on the sealed benchmark. No model parameters, thresholds, class weights, candidate-generation rules, or negative-sampling procedures were subsequently modified.

Final discriminative evaluation used the reference concept-text nodes as relation endpoints, thereby controlling for concept-identification error. For each of the 300 narratives, every ordered pair of distinct reference nodes was retained. No candidate pruning, category-based filtering, negative subsampling, or development-derived restriction was applied. Both frozen classifiers were therefore evaluated on the same 21,038 candidate pairs containing 248 positive relations and 20,790 \texttt{NO\_RELATION} instances.

The primary evaluation treated the task as three-class classification. An additional binary diagnostic analysis collapsed \texttt{RESULTS\_IN} and \texttt{CONTRIBUTES\_TO} into a single \texttt{RELATION} class to characterize relation detection independently of discrimination between the two positive relation types.

This gold-endpoint evaluation isolates candidate-level relation discrimination but does not measure the ability to discover safety concepts directly from raw narrative text. End-to-end extraction was therefore examined separately using the generative formulation described below.

\subsection{End-to-End Generative Evaluation}
\label{sec:generative_evaluation}

A complementary end-to-end evaluation analyzed previously generated outputs from \texttt{HuggingFaceTB/SmolLM2-1.7B-Instruct} for the same 300 narratives and gold relation set \cite{SmolLM2}. SmolLM2-1.7B was examined specifically as an ultra-lightweight, edge-deployable Small Language Model (SLM) to assess baseline zero- and few-shot relational extraction capabilities without task-specific fine-tuning. These outputs were generated in a separate experiment conducted using the same benchmark and were retained for complementary analysis in the present study. No additional SmolLM2 training, prompting, inference, or post-test optimization was performed for this analysis. Unlike the discriminative classifiers, SmolLM2 was not supplied with gold concept nodes or exhaustive candidate pairs. Instead, it received the narrative text and was required to identify relevant relation endpoints, infer directed relations, and return the extracted relations in a structured format. No task-specific fine-tuning was performed.

The outputs were generated under four prompting conditions: zero-shot, one-shot, few-shot, and retrieval-grounded prompting. The zero-shot condition contained no annotated demonstration, the one-shot condition used one fixed development example, and the few-shot condition used four fixed development examples. Retrieval-grounded prompting dynamically selected four demonstrations from a frozen pool of 12 annotated development examples using \texttt{sentence-transformers/all-MiniLM-L6-v2}. Narratives were represented using normalized 384-dimensional embeddings, and the four examples with the highest cosine similarity were retrieved for each target narrative.

The same SmolLM2 checkpoint and generation configuration were retained across the four conditions. Inference used 4-bit NormalFloat4 (NF4) quantization, FP16 computation, and double quantization. Generation was deterministic, with sampling disabled (\texttt{do\_sample=False}) and a maximum generation length of 1,400 new tokens. The zero-shot condition was processed sequentially, while the demonstration-based conditions used a batch size of eight for computational efficiency.

Generated relations were evaluated independently from concept extraction. A predicted relation was required to preserve the source endpoint, target endpoint, relation type, and direction of a reference relation. Semantic equivalence between generated and reference endpoints was permitted when supported by expert review. Candidate matches were independently assessed by three aviation-domain experts. Unanimous judgments were retained directly; cases in which two reviewers agreed and one disagreed were discussed to establish consensus, whereas matches supported by only one reviewer were rejected. Each generated and reference relation could participate in at most one accepted match.

Accepted relation matches were counted as true positives, unmatched generated relations as false positives, and unmatched reference relations as false negatives. Because generation relied on natural instruction prompting without constrained decoding or grammar masks (such as JSON schema enforcement or BNF grammars), relation labels outside the predefined \texttt{CONTRIBUTES\_TO} and \texttt{RESULTS\_IN} vocabulary and outputs that could not be reliably parsed into the required relation structure were retained as model errors rather than being corrected after generation, capturing realistic unconstrained generation behavior.

Because SmolLM2 jointly identified relation endpoints and generated relations, whereas the discriminative classifiers operated on explicitly enumerated pairs of gold concept endpoints, their numerical results were not treated as direct head-to-head model comparisons. The two evaluations instead provided complementary evidence about errors arising under end-to-end generation and decomposed relation classification.

\subsection{Evaluation Metrics and Diagnostic Analyses}
\label{sec:evaluation_metrics}

Three-class classifier performance was evaluated using accuracy, Macro-F1, weighted F1, and Matthews correlation coefficient (MCC). Accuracy quantified the overall proportion of correctly classified candidate pairs, while weighted F1 summarized class-specific F1 scores according to their empirical support. Because the exhaustive candidate space was strongly dominated by \texttt{NO\_RELATION}, these two measures were interpreted cautiously. Macro-F1 was emphasized as the primary class-balanced measure because it assigns equal importance to each relation class irrespective of class frequency. MCC was additionally reported as a summary measure derived from the complete multiclass confusion structure and was used to complement the F1-based measures under severe class imbalance\cite{Chicco2020}.

For binary relation detection, the two positive relation classes, \texttt{RESULTS\_IN} and \texttt{CONTRIBUTES\_TO}, were collapsed into a single \texttt{RELATION} class. True positives (TP) represented gold relations predicted as relational, false positives (FP) represented \texttt{NO\_RELATION} candidates incorrectly assigned a positive relation label, false negatives (FN) represented gold relations predicted as \texttt{NO\_RELATION}, and true negatives (TN) represented correctly rejected non-relational candidates. Precision, recall, and F1-score were calculated from these outcomes. Absolute TP, FP, FN, and TN counts were also retained because, under the naturally sparse candidate distribution, aggregate scores alone may obscure the scale of false-positive proliferation.

For the end-to-end SmolLM2 evaluation, relation-level precision, recall, and F1-score were calculated from accepted relation matches, unsupported generated relations, and unrecovered gold relations. Accepted matches were treated as true positives, unmatched generated relations as false positives, and unrecovered gold relations as false negatives. True-negative counts were not defined for this setting because the generative model did not classify an exhaustively enumerated set of negative candidate pairs. SmolLM2 metrics were therefore interpreted within the end-to-end generative formulation rather than as directly equivalent to the candidate-level discriminative metrics.

Post-test diagnostic analyses were conducted to characterize the error structure without modifying the frozen models. For RoBERTa, relation confidence was defined as

\begin{equation}
p_{\mathrm{rel}}
=
p(\texttt{RESULTS\_IN})
+
p(\texttt{CONTRIBUTES\_TO}),
\end{equation}

and examined separately across true-positive, false-positive, false-negative, and true-negative candidates. Distributional summaries included the number of candidates, mean, median, first and third quartiles, minimum, and maximum relation probability for each outcome group. These analyses were used to examine whether correctly and incorrectly classified candidates occupied distinguishable confidence ranges. No post-test confidence threshold was selected or applied.

Errors were additionally analyzed by directed source--target concept-category combination. Because consolidated concept-text nodes could retain multiple category memberships, this analysis was conducted at the category-membership level rather than by treating category-direction counts as mutually exclusive candidate pairs. Category annotations were used only after prediction for descriptive error analysis and were not supplied to either classifier as predictive features or used to filter candidates.

Cross-model agreement between the SVM and RoBERTa classifiers was examined at both the three-class and binary relation/no-relation levels using the proportion of identical predictions and Cohen's $\kappa$\cite{Cohen1960}. Candidate pairs predicted as relational by both classifiers were additionally compared with the gold relation set to distinguish shared correct predictions from shared false positives. Model-specific positive predictions were also examined to characterize differences in relation-prediction behavior between the two classifiers.

All post-test confidence, category-direction, and cross-model agreement analyses were diagnostic. They did not alter model parameters, class weights, decision thresholds, candidate construction, model selection, or the predictions used for the primary sealed evaluation.

\section{Results}
\label{sec:results}

\subsection{Development-Set Model Selection}
\label{sec:results_development}

Table~\ref{tab:development_results} summarizes the performance of the TF--IDF + Linear SVM and weighted RoBERTa classifiers on the 174-instance development validation set. The models exhibited different performance profiles: the SVM achieved higher accuracy and weighted F1, whereas RoBERTa achieved higher Macro-F1 and MCC.

\begin{table}[H]
\caption{Performance of the relation classifiers on the frozen development validation set. Macro-F1 was the predefined primary model-selection criterion. Best values for each metric are shown in bold.\label{tab:development_results}}
\begin{tabularx}{\textwidth}{LCCCCCCC}
\toprule
\textbf{Model} &
\textbf{Accuracy} &
\textbf{Macro-F1} &
\textbf{Weighted F1} &
\textbf{MCC} &
\textbf{NO\_REL. F1} &
\textbf{RESULTS\_IN F1} &
\textbf{CONTRIB. F1} \\
\midrule
TF--IDF + Linear SVM &
\textbf{0.5460} &
0.4738 &
\textbf{0.5605} &
0.1941 &
\textbf{0.6445} &
0.5000 &
0.2769 \\
Weighted RoBERTa &
0.5000 &
\textbf{0.4809} &
0.5143 &
\textbf{0.2447} &
0.5464 &
\textbf{0.5060} &
\textbf{0.3902} \\
\bottomrule
\end{tabularx}
\end{table}

The SVM achieved a Macro-F1 of 0.4738 and an MCC of 0.1941, compared with 0.4809 and 0.2447, respectively, for RoBERTa. At the class level, RoBERTa achieved higher F1 scores for both positive relation classes: 0.5060 for \texttt{RESULTS\_IN} and 0.3902 for \texttt{CONTRIBUTES\_TO}, compared with 0.5000 and 0.2769 for the SVM. Conversely, the SVM performed better on \texttt{NO\_RELATION}, with an F1 of 0.6445 compared with 0.5464 for RoBERTa.

The validation confusion matrices further illustrate this difference. With rows representing reference classes and columns representing predicted classes in the order \texttt{NO\_RELATION}, \texttt{RESULTS\_IN}, and \texttt{CONTRIBUTES\_TO}, the SVM confusion matrix was

\begin{equation}
\mathbf{C}_{\mathrm{SVM}} =
\begin{bmatrix}
68 & 24 & 24 \\
8 & 18 & 4 \\
19 & 0 & 9
\end{bmatrix},
\end{equation}

whereas the RoBERTa confusion matrix was

\begin{equation}
\mathbf{C}_{\mathrm{RoBERTa}} =
\begin{bmatrix}
50 & 29 & 37 \\
8 & 21 & 1 \\
9 & 3 & 16
\end{bmatrix}.
\end{equation}

RoBERTa correctly classified 21 of the 30 validation \texttt{RESULTS\_IN} instances and 16 of the 28 \texttt{CONTRIBUTES\_TO} instances, compared with 18 and 9, respectively, for the SVM. This greater positive-class recovery was accompanied by more \texttt{NO\_RELATION} instances being assigned a positive relation label.

Because Macro-F1 was the predefined primary selection criterion, weighted RoBERTa was selected as the primary discriminative classifier. The SVM was retained as the classical baseline. All model configurations were frozen before evaluation on the sealed benchmark.

\subsection{Sealed Gold-Endpoint Relation Classification}
\label{sec:results_sealed}

The frozen classifiers were subsequently evaluated on the sealed benchmark containing all 21,038 directed pairs generated from 2,524 unique reference concept-text nodes. The candidate space comprised 20,790 \texttt{NO\_RELATION} instances, 228 \texttt{RESULTS\_IN} instances, and 20 \texttt{CONTRIBUTES\_TO} instances. Positive relations therefore represented only 1.18\% of the exhaustive candidate space.

Table~\ref{tab:sealed_results} reports the resulting three-class performance. The SVM achieved higher accuracy, Macro-F1, weighted F1, and \texttt{NO\_RELATION} F1, whereas RoBERTa achieved slightly higher MCC and higher F1 for both positive relation classes.

\begin{table}[H]
\caption{Performance of the frozen relation classifiers on the sealed exhaustive gold-endpoint candidate set. The benchmark contained 21,038 ordered candidate pairs, of which 248 (1.18\%) represented positive relations. Bold values indicate the higher value between the two models for each metric and do not represent post-test model selection.\label{tab:sealed_results}}
\begin{tabularx}{\textwidth}{LCCCCCCC}
\toprule
\textbf{Model} &
\textbf{Accuracy} &
\textbf{Macro-F1} &
\textbf{Weighted F1} &
\textbf{MCC} &
\textbf{NO\_REL. F1} &
\textbf{RESULTS\_IN F1} &
\textbf{CONTRIB. F1} \\
\midrule
TF--IDF + Linear SVM &
\textbf{0.6582} &
\textbf{0.2741} &
\textbf{0.7844} &
0.0128 &
\textbf{0.7935} &
0.0268 &
0.0021 \\
Weighted RoBERTa &
0.5262 &
0.2420 &
0.6802 &
\textbf{0.0197} &
0.6880 &
\textbf{0.0331} &
\textbf{0.0049} \\
\bottomrule
\end{tabularx}
\end{table}

The moderate accuracy and weighted F1 values were strongly influenced by the dominant \texttt{NO\_RELATION} class. In contrast, Macro-F1 decreased to 0.2741 for the SVM and 0.2420 for RoBERTa, while MCC decreased to 0.0128 and 0.0197, respectively. Performance on the positive classes was particularly low: \texttt{RESULTS\_IN} F1 was 0.0268 for the SVM and 0.0331 for RoBERTa, while \texttt{CONTRIBUTES\_TO} F1 was 0.0021 and 0.0049, respectively. Because \texttt{CONTRIBUTES\_TO} was extremely scarce ($n=20$, representing only 0.095\% of candidate pairs), its near-zero F1 substantially deflated the overall three-class Macro-F1. This acute imbalance motivates the complementary binary relation-detection evaluation in Section~\ref{sec:relation_detection}, which isolates overall relation discovery from fine-grained discrimination between the two positive relation types.

The sealed-test confusion matrices were

\begin{equation}
\mathbf{C}_{\mathrm{SVM}}^{\mathrm{test}} =
\begin{bmatrix}
13772 & 5140 & 1878 \\
134 & 73 & 21 \\
18 & 0 & 2
\end{bmatrix},
\end{equation}

and

\begin{equation}
\mathbf{C}_{\mathrm{RoBERTa}}^{\mathrm{test}} =
\begin{bmatrix}
10966 & 5361 & 4463 \\
115 & 94 & 19 \\
8 & 1 & 11
\end{bmatrix}.
\end{equation}

RoBERTa correctly classified 94 of the 228 \texttt{RESULTS\_IN} instances and 11 of the 20 \texttt{CONTRIBUTES\_TO} instances, compared with 73 and 2, respectively, for the SVM. However, both classifiers also assigned positive relation labels to large numbers of reference \texttt{NO\_RELATION} candidates. RoBERTa classified 5,361 negative candidates as \texttt{RESULTS\_IN} and 4,463 as \texttt{CONTRIBUTES\_TO}, while the corresponding SVM counts were 5,140 and 1,878.

Thus, the sealed evaluation produced a markedly different performance profile from the sampled development setting. Both models continued to recover reference relations when supplied with gold concept endpoints, but their positive predictions were accompanied by substantial numbers of false positives.

\subsection{Relation Detection under Natural Candidate Sparsity}
\label{sec:relation_detection}

To characterize relation detection independently of the distinction between the two positive relation types, the three-class predictions were collapsed into a binary relation/no-relation representation. The 248 \texttt{RESULTS\_IN} and \texttt{CONTRIBUTES\_TO} instances formed the positive class, while the remaining 20,790 candidate pairs formed the negative class.

As shown in Table~\ref{tab:binary_relation_results}, the SVM recovered 96 of the 248 reference relations and missed 152, corresponding to a recall of 0.3871. Its 7,114 positive predictions included 7,018 false positives, resulting in a precision of 0.0135 and an F1 of 0.0261. RoBERTa recovered 125 reference relations and missed 123, yielding a recall of 0.5040. However, 9,824 of its 9,949 positive predictions were false positives, resulting in a precision of 0.0126 and an F1 of 0.0245.

\begin{table}[H]
\caption{Binary relation-detection results on the sealed exhaustive candidate set. The two positive relation classes were collapsed into a single relation-positive class for this diagnostic analysis.\label{tab:binary_relation_results}}
\begin{tabularx}{\textwidth}{LCCCCCCC}
\toprule
\textbf{Model} &
\textbf{TP} &
\textbf{FP} &
\textbf{FN} &
\textbf{TN} &
\textbf{Precision} &
\textbf{Recall} &
\textbf{F1} \\
\midrule
TF--IDF + Linear SVM &
96 &
7018 &
152 &
13772 &
0.0135 &
0.3871 &
0.0261 \\
Weighted RoBERTa &
125 &
9824 &
123 &
10966 &
0.0126 &
0.5040 &
0.0245 \\
\bottomrule
\end{tabularx}
\end{table}

False-positive proliferation was therefore the dominant error pattern. RoBERTa generated approximately 79 false-positive predictions for every true-positive prediction, while the SVM generated approximately 73. RoBERTa's higher relation recall was consequently accompanied by a substantially larger false-positive burden.

\begin{figure}[H]
\centering
\includegraphics[width=\textwidth]{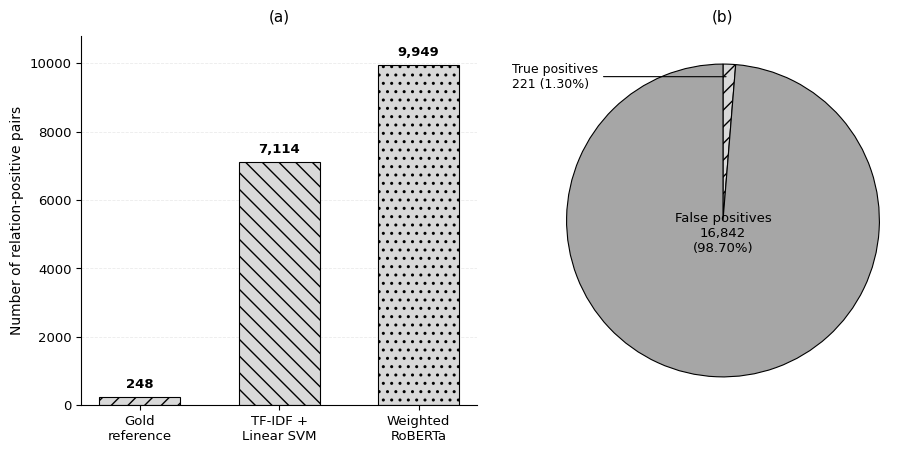}
\caption{\textbf{Relation-prediction proliferation under exhaustive candidate generation.}
(a) \textbf{Relation-positive outputs:} Number of reference relations compared with the total number of relation-positive predictions produced by the frozen TF--IDF + Linear SVM and weighted RoBERTa classifiers.
(b) \textbf{Composition of predicted relations:} Composition of each model's relation-positive predictions as true positives and false positives.
The sealed benchmark contained 21,038 ordered candidate pairs, of which only 248 (1.18\%) were positive reference relations.}
\label{fig:relation_prediction_proliferation}
\end{figure}

Figure~\ref{fig:relation_prediction_proliferation} illustrates the scale and composition of the relation-positive outputs. Against 248 reference relations, the SVM produced 7,114 relation-positive predictions and RoBERTa produced 9,949, corresponding to approximately 28.69 and 40.12 times the number of reference relations, respectively. True positives represented 1.35\% of the SVM's positive predictions and 1.26\% of RoBERTa's.

\unskip

\begin{figure}[H]
\centering
\includegraphics[width=\textwidth]{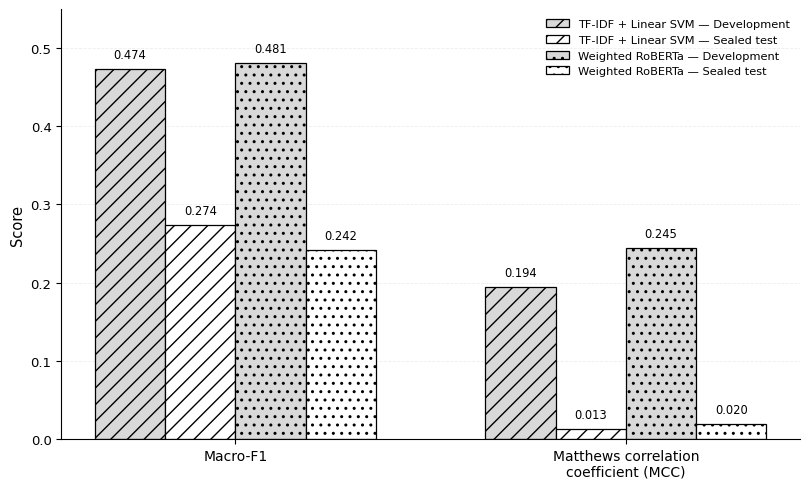}
\caption{\textbf{Performance transition from sampled development validation to exhaustive sealed evaluation.}
Macro-F1 and Matthews correlation coefficient (MCC) are shown for the TF--IDF + Linear SVM and weighted RoBERTa classifiers. Development evaluation used the sampled hard-negative distribution (2:1 negative-to-positive ratio), whereas sealed evaluation used exhaustive ordered candidate pairs, of which 98.82\% were \texttt{NO\_RELATION}. Both models exhibited substantially lower Macro-F1 and MCC under the naturally sparse exhaustive candidate distribution.}
\label{fig:generalization}
\end{figure}

The transition from the sampled development distribution to the exhaustive sealed candidate space was also accompanied by substantial reductions in Macro-F1 and MCC (Figure~\ref{fig:generalization}). SVM Macro-F1 decreased from 0.4738 to 0.2741 and MCC from 0.1941 to 0.0128. RoBERTa Macro-F1 decreased from 0.4809 to 0.2420 and MCC from 0.2447 to 0.0197.

\unskip

These results show that performance under the controlled hard-negative development distribution did not transfer directly to exhaustive candidate evaluation. Despite access to the gold concept endpoints, both models showed limited discrimination between the small number of genuine relations and the much larger set of non-relational candidate pairs.

\subsection{RoBERTa Confidence and Error Analysis}
\label{sec:confidence_analysis}

To further characterize RoBERTa's false-positive behavior, relation probabilities were examined according to binary prediction outcome. For each candidate pair, relation probability was defined as the sum of the probabilities assigned to \texttt{RESULTS\_IN} and \texttt{CONTRIBUTES\_TO}. Table~\ref{tab:confidence_outcomes} summarizes the distributions for true positives (TP), false positives (FP), false negatives (FN), and true negatives (TN).

\begin{table}[H]
\caption{Distribution of RoBERTa relation probabilities according to binary sealed-test prediction outcome. Relation probability is defined as $p_{\mathrm{rel}}=p(\texttt{RESULTS\_IN})+p(\texttt{CONTRIBUTES\_TO})$.\label{tab:confidence_outcomes}}

\begin{tabularx}{\textwidth}{
    >{\raggedright\arraybackslash}X
    *{7}{>{\centering\arraybackslash}p{1.25cm}}
}
\toprule
\textbf{Outcome} &
\textbf{$n$} &
\textbf{Mean} &
\textbf{Median} &
\textbf{Q1} &
\textbf{Q3} &
\textbf{Minimum} &
\textbf{Maximum} \\
\midrule
True positive  & 125   & 0.6787 & 0.7002 & 0.6234 & 0.7348 & 0.5168 & 0.7928 \\
False positive & 9824  & 0.6555 & 0.6602 & 0.5973 & 0.7153 & 0.5149 & 0.8001 \\
False negative & 123   & 0.3671 & 0.3504 & 0.3073 & 0.4221 & 0.2501 & 0.5220 \\
True negative  & 10966 & 0.3800 & 0.3727 & 0.3184 & 0.4416 & 0.2186 & 0.5578 \\
\bottomrule
\end{tabularx}
\vspace{-8pt}
\end{table}

The true-positive and false-positive distributions overlapped substantially. True positives had a mean relation probability of 0.6787 and a median of 0.7002, compared with 0.6555 and 0.6602, respectively, for false positives. Their interquartile ranges were similarly overlapping: 0.6234--0.7348 for true positives and 0.5973--0.7153 for false positives.

Substantial overlap was also observed among relation-negative predictions. False negatives had a mean relation probability of 0.3671 and a median of 0.3504, compared with 0.3800 and 0.3727 for true negatives. The corresponding interquartile ranges were 0.3073--0.4221 and 0.3184--0.4416.

\begin{figure}[H]
\centering
\includegraphics[width=\textwidth]{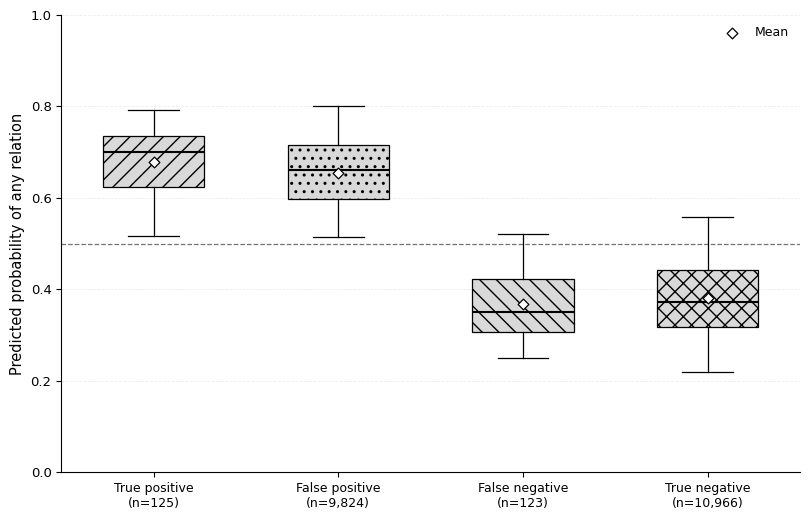}
\caption{\textbf{Distribution of frozen RoBERTa relation probabilities on the sealed benchmark.}
Distributions are shown for true-positive (TP), false-positive (FP), false-negative (FN), and true-negative (TN) candidate pairs. Relation probability was calculated as $p(\texttt{RESULTS\_IN}) + p(\texttt{CONTRIBUTES\_TO})$. The dashed horizontal line at 0.5 provides a visual reference for the binary relation/no-relation interpretation of the frozen three-class outputs and was not selected through post-test threshold optimization.}
\label{fig:confidence_distribution}
\end{figure}

Figure~\ref{fig:confidence_distribution} visualizes these distributions. The similarity between the TP and FP distributions is particularly consequential because the false-positive group contained 9,824 candidates, compared with only 125 true positives. Thus, overlapping score distributions produced a substantial absolute false-positive burden under the sparse candidate distribution.

\unskip

The observed overlap shows that the false positives were not confined to a narrow set of marginal predictions near a decision boundary. Many false-positive candidates received relation probabilities within the same range as correctly detected reference relations. Similarly, the FN and TN distributions showed no clear separation. No post-test threshold adjustment was therefore performed; the confidence analysis was retained solely as a diagnostic of the frozen model.

\subsection{Structural Distribution of False Positives and Cross-Model Agreement}
\label{sec:structural_errors}

Post-test diagnostic analysis examined whether false-positive predictions were concentrated in particular source--target category directions. Because consolidated concept nodes could retain multiple category memberships, the 21,038 ordered candidate pairs corresponded to 29,094 category-direction memberships across 64 source--target category combinations. These membership counts were used only for descriptive analysis and did not alter the candidate set or model predictions.

Table~\ref{tab:category_errors} reports selected high-frequency category directions with substantial false-positive counts. The largest concentration occurred for \textit{event}$\rightarrow$\textit{event} pairs, comprising 3,570 candidate memberships and 151 positive reference-relation memberships. RoBERTa recovered 84 positive memberships but produced 1,783 false-positive memberships, while the SVM produced 1,268 false-positive memberships.

\begin{table}[H]
\caption{Selected source--target concept-category directions from the sealed-test diagnostic analysis. TP and FP denote true-positive and false-positive relation predictions, respectively. Counts are category-membership based; because concept nodes may belong to multiple categories, rows are not mutually exclusive and should not be summed to recover the 21,038 unique candidate pairs.\label{tab:category_errors}}
\begin{tabularx}{\textwidth}{
    >{\raggedright\arraybackslash}X
    >{\centering\arraybackslash}p{1.45cm}
    >{\centering\arraybackslash}p{1.25cm}
    >{\centering\arraybackslash}p{1.35cm}
    >{\centering\arraybackslash}p{1.35cm}
    >{\centering\arraybackslash}p{1.10cm}
}
\toprule
\textbf{Source $\rightarrow$ Target} &
\textbf{Candidates} &
\textbf{Ref. Pos.} &
\textbf{RoBERTa TP} &
\textbf{RoBERTa FP} &
\textbf{SVM FP} \\
\midrule
Event $\rightarrow$ Event & 3570 & 151 & 84 & 1783 & 1268 \\
Event $\rightarrow$ Phase of Flight & 1859 & 39 & 34 & 975 & 610 \\
Phase of Flight $\rightarrow$ Event & 1859 & 4 & 4 & 943 & 778 \\
Consequence $\rightarrow$ Event & 1503 & 9 & 5 & 938 & 573 \\
Hazard $\rightarrow$ Event & 2090 & 52 & 33 & 909 & 608 \\
Event $\rightarrow$ Hazard & 2090 & 5 & 2 & 885 & 760 \\
Event $\rightarrow$ Consequence & 1503 & 61 & 31 & 857 & 659 \\
Environmental Factor $\rightarrow$ Event & 526 & 19 & 16 & 439 & 163 \\
\bottomrule
\end{tabularx}

\end{table}

False positives were not confined to category directions without reference relations. For example, \textit{event}$\rightarrow$\textit{consequence}, \textit{hazard}$\rightarrow$\textit{event}, and \textit{environmental factor}$\rightarrow$\textit{event} contained genuine relations while also accumulating substantial numbers of false-positive predictions. Conversely, directions with few positive reference relations could also generate numerous false positives. The observed pattern therefore reflects difficulty discriminating individual candidate pairs within structurally plausible relation spaces rather than prediction only across obviously incompatible category combinations.

Restricting inference retrospectively to category directions observed among positive benchmark relations could remove some false-positive candidates, but doing so would use information derived from the sealed reference labels. Moreover, because genuine and false relations frequently occurred within the same category directions, category-level filtering alone would not resolve candidate-level discrimination. No post-test category filtering was applied.

Cross-model agreement analysis further showed that the false-positive behavior was not confined to one model family. Across the 21,038 candidates, the SVM and RoBERTa produced the same three-class prediction for 11,217 instances, corresponding to an exact agreement of 0.5332 and Cohen's $\kappa$ of 0.1775. After predictions were collapsed into relation-positive and relation-negative classes, the models agreed on 12,123 candidates (0.5762), with Cohen's $\kappa=0.1374$.

Both classifiers predicted a positive relation for 4,074 candidate pairs. Only 63 of these shared positive predictions corresponded to reference relations, while 4,011 were false positives. RoBERTa produced an additional 5,875 relation-positive predictions not shared by the SVM, whereas the SVM produced 3,040 additional positive predictions not shared by RoBERTa.

Thus, although the two model families did not make identical candidate-level decisions, substantial false-positive proliferation occurred in both. Even the intersection of their positive predictions remained dominated by false positives. Cross-model agreement was therefore retained as a diagnostic analysis and was not used retrospectively to construct an ensemble.

\subsection{End-to-End SmolLM2 Relation Extraction}
\label{sec:results_smollm}

The complementary end-to-end evaluation examined previously generated SmolLM2-1.7B-Instruct outputs for the same 300 benchmark narratives and 248 reference relations. Unlike the gold-endpoint classifiers, SmolLM2 was required to identify relation endpoints and generate directed relations directly from each narrative. Four prompting conditions were evaluated: zero-shot, one-shot, few-shot, and retrieval-grounded prompting.

Table~\ref{tab:smollm_relation_results} reports the relation-level results. Zero-shot prompting recovered no reference relations. One-shot prompting recovered 15 relations, with a precision of 0.0534, recall of 0.0605, and F1 of 0.0567. Few-shot prompting recovered 9 relations, yielding an F1 of 0.0182, while retrieval-grounded prompting recovered 13 relations and achieved an F1 of 0.0289.

\begin{table}[H]
\caption{End-to-end SmolLM2-1.7B-Instruct relation-extraction results on the 300-narrative benchmark. Each prompting condition was evaluated against the same 248 reference relations. TP, FP, and FN denote relation-level true positives, false positives, and false negatives, respectively.\label{tab:smollm_relation_results}}
\begin{tabularx}{\textwidth}{LCCCCCC}
\toprule
\textbf{Prompting Condition} &
\textbf{TP} &
\textbf{FP} &
\textbf{FN} &
\textbf{Precision} &
\textbf{Recall} &
\textbf{F1} \\
\midrule
Zero-shot &
0 & 129 & 248 &
0.0000 & 0.0000 & 0.0000 \\
One-shot &
15 & 266 & 233 &
0.0534 & 0.0605 & 0.0567 \\
Few-shot &
9 & 734 & 239 &
0.0121 & 0.0363 & 0.0182 \\
Retrieval-grounded &
13 & 638 & 235 &
0.0200 & 0.0524 & 0.0289 \\
\bottomrule
\end{tabularx}
\end{table}

Across the four prompting conditions, each of the 248 benchmark relations was evaluated once per condition, yielding 992 condition-specific reference-relation instances. SmolLM2 recovered 37 of these instances, generated 1,767 false-positive relations, and missed 955, corresponding to an aggregated precision of 0.0205, recall of 0.0373, and F1 of 0.0265.

The structural audit showed that false-positive outputs included both structurally valid but unsupported relations and malformed outputs. Of the 734 few-shot false positives, 548 were structurally valid and 186 were malformed. The corresponding counts were 202 and 64 under one-shot prompting and 281 and 357 under retrieval-grounded prompting. All 129 zero-shot false positives were classified as malformed.

All 37 matched relation instances were established through the semantic matching procedure rather than exact relation-string matching. One-shot, few-shot, and retrieval-grounded prompting produced 15, 9, and 13 semantic matches, respectively, whereas zero-shot prompting produced none.

The SmolLM2 and gold-endpoint experiments represent different prediction tasks and are therefore not interpreted as a direct numerical comparison of model performance. SmolLM2 had to identify concepts, determine source--target structure, generate a relation, and produce a valid structured output. The discriminative classifiers were instead supplied with gold concept endpoints but were required to classify every ordered candidate pair in the exhaustive candidate space.

Nevertheless, the two settings revealed complementary error profiles. The end-to-end generative setting showed low recovery of reference relations together with unsupported and malformed outputs, whereas the exhaustive gold-endpoint setting recovered a larger proportion of reference relations but produced substantial false-positive proliferation. The implications of these findings are examined in the Discussion.

\section{Discussion}
\label{sec:discussion}

\subsection{Principal Findings and Relational Sparsity}
\label{sec:discussion_principal}

This study examined directed relation extraction from aviation accident narratives using a decomposed framework that separated relation discrimination from concept identification. By supplying gold concept endpoints to the discriminative classifiers, the evaluation isolated whether a model could determine which ordered concept pairs expressed \texttt{RESULTS\_IN}, \texttt{CONTRIBUTES\_TO}, or \texttt{NO\_RELATION}. The central finding is that access to the correct concept endpoints was not sufficient for precise relation extraction when classification was extended to the complete candidate space.

This limitation became apparent when the models moved from controlled hard-negative development to exhaustive sealed evaluation. Both the TF--IDF + Linear SVM and weighted RoBERTa learned useful relational signals during development, but their Macro-F1 and MCC decreased substantially when positive relations represented only 1.18\% of the sealed candidate space. The consistency of this degradation across a sparse linear classifier and a pretrained Transformer suggests that the difficulty was not confined to a single model family. These findings address RQ1 and RQ2 by showing that relation discrimination was feasible under controlled development conditions but did not remain sufficiently precise under the naturally sparse exhaustive candidate distribution.

The dominant problem was false-positive proliferation rather than complete failure to recover genuine relations. RoBERTa recovered 125 of the 248 reference relations and the SVM recovered 96, but these predictions were accompanied by 9,824 and 7,018 false positives, respectively. Thus, recovering a meaningful proportion of positive relations did not translate into a precise relational representation. This distinction is important in sparse relation extraction because even a modest tendency to assign positive labels to negative candidates can generate many unsupported relations when the candidate space is large.

The post-test diagnostics further showed that the problem could not be explained solely by obviously incompatible concept combinations. False positives occurred extensively within source--target category directions that also contained genuine relations, and RoBERTa's confidence distributions showed substantial overlap between true-positive and false-positive candidates. Moreover, only 63 of the 4,074 candidate pairs predicted as relational by both classifiers corresponded to reference relations. These findings address RQ3 by indicating that the principal difficulty lies in discriminating the relatively few supported relations from many structurally and semantically plausible alternatives.

The results also clarify the role of negative sampling in relation-extraction evaluation. Negative sampling remains useful during model development because it provides informative negative examples without requiring exhaustive enumeration of the candidate space, while hard-negative strategies can focus learning on particularly difficult negative instances \cite{Xu2022NegativeSampling,Olasunkanmi2025RELATE}. However, sampled-negative validation and exhaustive evaluation answer different questions. The former assesses whether a model can learn relational distinctions under a controlled distribution, whereas the latter reveals the cumulative consequences of classification errors when all candidate pairs are considered. In the present study, the substantial performance decrease between these settings demonstrates the importance of evaluating relation-extraction systems under the candidate distribution relevant to their intended application.

Candidate generation should therefore be treated as an explicit component of the extraction problem rather than as a neutral preprocessing step. In operational knowledge-graph construction, this challenge can be addressed through a multi-stage architecture. First, candidate-proposal mechanisms based on sentence-boundary co-occurrence, syntactic dependency paths, or shallow semantic distance can prune the unconstrained quadratic candidate space ($n(n-1)$) to a manageable subset of plausible local and cross-clause hypotheses before computationally intensive relation classification. Second, candidate relations proposed by discriminative models can be routed through a verification stage grounded in Retrieval-Augmented Generation (RAG) or Natural Language Inference (NLI). In such a framework, candidate entity pairs serve as retrieval queries to pinpoint specific narrative evidence passages, which an NLI verifier or structured verification prompt evaluates for explicit textual entailment before committing directed edges to a safety knowledge graph. Such mechanisms would need to be developed without access to final benchmark labels and evaluated for both the unsupported candidates they remove and the genuine relations they retain. The present results therefore do not argue against hard-negative sampling; rather, they show that development under sampled negatives should be complemented by evaluation under the sparse candidate distribution relevant to the intended extraction setting.

\subsection{Discriminative and End-to-End Generative Relation Extraction}
\label{sec:discussion_discriminative_generative}

The discriminative and generative experiments examined different stages of structured relation extraction and should therefore be interpreted as complementary rather than directly comparable evaluations. In the decomposed setting, the SVM and RoBERTa classifiers received gold concept endpoints and were responsible only for relation discrimination. SmolLM2 operated end-to-end and had to identify relevant concepts, determine source--target structure, assign relation semantics, and produce a valid structured output.

The gold-endpoint experiment is diagnostically important because it removes concept discovery as an explanation for relation-extraction errors. Its results demonstrate that relation discrimination remains difficult even when the correct concepts are already available. The end-to-end generative evaluation exposed additional sources of error, including missed relations, unsupported relations, and malformed outputs. Similar challenges are relevant to recent aviation research using generative, retrieval-augmented, and knowledge-graph approaches for extracting structured information from safety narratives \cite{Ren2025RACI,Xiang2026RAGKG,Chen2024AviationKG,Cheng2026ASHCKG}.

The prompting conditions also showed that providing additional demonstrations or retrieved context did not produce a consistent improvement for the evaluated SmolLM2 configuration. One-shot prompting produced the strongest relation-level F1 among the four conditions, whereas few-shot and retrieval-grounded prompting generated larger numbers of unsupported outputs. This observation is specific to the model, prompts, retrieval configuration, ontology, and benchmark used in this study and should not be interpreted as a general limitation of few-shot prompting or retrieval-augmented generation.

The generative evaluation additionally highlights an evaluation challenge associated with free-form output. All matched SmolLM2 relation instances were identified through the predefined semantic-matching procedure rather than exact relation-string correspondence. Generated expressions may therefore preserve relevant meaning without reproducing the lexical form of the reference annotation. At the same time, semantic matching requires consistent adjudication to determine whether a generated relation represents the intended reference structure.

Together, the two experiments address RQ4 by locating errors at different stages of the extraction process. End-to-end evaluation incorporates concept identification, endpoint pairing, relation assignment, directionality, and structured-output generation, whereas gold-endpoint evaluation isolates relation discrimination. This decomposition provides a more informative account of system behavior than a single aggregate relation score because it helps identify where errors enter the extraction pipeline.

These complementary error profiles also motivate investigation of hybrid architectures. Future systems could combine concept identification with learned candidate proposal, relation classification, and constrained structured generation or verification. Such designs could reduce exhaustive pair enumeration while limiting unsupported generative outputs. Furthermore, incorporating grammar-constrained decoding (such as JSON schema enforcement or regular-expression guided decoding) within medium-sized open instruction-tuned language models (e.g., 7B- to 14B-parameter architectures) represents an immediate avenue to eliminate syntactic malformations and unparseable outputs, thereby isolating genuine semantic extraction and factual hallucination rates. Their effectiveness, however, would need to be established through independent development and evaluation rather than through post-test filtering of the present benchmark.

\subsection{Implications for Aviation Safety Knowledge Discovery}
\label{sec:discussion_implications}

Reliable relation extraction could support structured representations of aviation accident narratives in which analysts examine not only which hazards, technical factors, environmental conditions, human factors, events, and consequences are present, but also how these concepts are connected. Recent aviation research has demonstrated the potential of causal extraction and knowledge-graph approaches for organizing and reasoning over safety information \cite{Chen2026ARCEM,Ren2025RACI,Chen2024AviationKG,Xiang2026RAGKG,Cheng2026ASHCKG,Lin2026AI4AirSafe}. The present findings emphasize that the reliability of extracted relations is critical to such applications.

In a knowledge graph or other structured representation, a false-positive relation becomes an unsupported edge between otherwise valid concepts. If such predictions accumulate, they can distort the relational structure of the source narrative and potentially affect downstream tasks that depend on graph connectivity, including relational retrieval, identification of recurring safety-factor combinations, and analysis of accident mechanisms. Accurate concept identification should therefore not be treated as sufficient evidence that a downstream relation-extraction component will produce an equally reliable relational representation.

The findings support a staged view of automated aviation safety knowledge extraction in which concept identification, candidate proposal, relation classification, and structured representation are treated as distinguishable components. Candidate proposal deserves particular attention because exhaustive enumeration provides a stringent evaluation of relation discrimination but also exposes the classifier to a rapidly expanding set of non-relational candidates. A learned proposal mechanism could reduce this search space, provided that its ability to retain genuine relations is evaluated alongside its ability to remove unsupported candidates.

Provenance and uncertainty information may also improve the transparency of automatically constructed safety representations. Future systems could retain the originating narrative, participating concept spans, supporting textual evidence, model confidence, and extraction method alongside each predicted relation. These attributes could support verification or analyst review before extracted relations are incorporated into higher-level knowledge resources. The present study does not establish the effectiveness of such mechanisms, but its findings provide a clear motivation for evaluating them.

The results also demonstrate why evaluation measures should reflect the intended use of relational outputs. In a candidate space dominated by \texttt{NO\_RELATION}, accuracy and weighted F1 can obscure poor performance on the comparatively rare relations of primary interest. Macro-F1, MCC, relation-specific precision and recall, absolute false-positive counts, and evaluation over the complete candidate space provide complementary information about whether an extraction system can preserve the distinction between supported and unsupported connections at scale.

\subsection{Limitations and Future Research}
\label{sec:discussion_limitations}

Several limitations should be considered when interpreting these findings. First, the sealed benchmark contained 300 accident narratives and 248 positive reference relations. Although exhaustive enumeration provided a large candidate space for evaluating relation discrimination, the number of positive relations remained limited, particularly for \texttt{CONTRIBUTES\_TO}, which was represented by only 20 instances. The very low performance for this class should therefore be interpreted cautiously, and larger relation-annotated datasets are needed to assess less frequently represented relational semantics.

Second, the primary discriminative experiment intentionally used gold concept endpoints. This design isolated relation classification from concept-identification errors but does not represent a complete operational extraction pipeline. An end-to-end system would additionally depend on the quality of automatically identified concepts. Future studies should evaluate concept detection and relation classification jointly while retaining a decomposed evaluation that allows errors at different stages to be distinguished.

Third, the classifiers were developed using an independent graph-annotated resource whose ontology required conservative harmonization with the target benchmark. Only relations that could be mapped without changing their intended semantics were retained. This reduced the available development data but avoided imposing unsupported equivalences between annotation schemes. The substantial difference between the sampled development distribution and the exhaustive sealed distribution also means that development performance did not directly characterize behavior under natural candidate sparsity.

Fourth, the discriminative evaluation considered only a TF--IDF Linear SVM and RoBERTa-based classifier. Their shared false-positive behavior suggests that the observed problem was not unique to one architecture, but the study does not establish how other relation-extraction methods would behave under the same protocol. Future comparisons could include span-aware relation models, joint entity--relation architectures, structured prediction methods, relation-proposal networks, and larger pretrained language models.

The confidence, category-direction, and cross-model agreement analyses were post-test diagnostics and were not used to modify thresholds, candidate rules, class weights, model parameters, or model selection. The study therefore does not determine whether independently developed calibration, candidate ranking, or thresholding methods could improve precision. Such methods should be developed on separate training and validation resources before evaluation on a sealed benchmark.

An evidence-verification component was also explored during development but was excluded before sealed evaluation because its development performance did not support inclusion in the final framework. The verifier achieved a Macro-F1 of 0.4834, an MCC of 0.007, and an F1 of 0.1039 for the supported-relation class. Evidence-grounded verification nevertheless remains a potential research direction if suitable relation-support annotations can be developed independently.

The SmolLM2 findings are similarly limited to the evaluated model and prompting configurations. They do not characterize generative language models in general. Larger models, relation-specific prompting, constrained decoding, structured-output mechanisms, and alternative retrieval strategies may produce different results and warrant independent investigation.

Finally, the benchmark was derived from NTSB accident narratives and the relation ontology used in this study. Generalization to other aviation reporting systems, narrative styles, time periods, or safety ontologies has not been established. External validation on independently annotated corpora is therefore necessary to determine whether the observed relational-sparsity and false-positive patterns persist across reporting environments.

Future research should consequently prioritize larger and more diverse relation-annotated resources, independently trained candidate-proposal mechanisms, evidence-grounded verification, and joint or constrained extraction architectures. Importantly, methodological improvements should continue to be evaluated using strictly separated development and test resources and under candidate distributions that preserve the relational sparsity encountered when complete narratives are processed.

\section{Conclusions}
\label{sec:conclusions}

This study investigated directed relation extraction from aviation accident narratives using a decomposed framework that separated relation discrimination from concept identification. By supplying gold concept endpoints and evaluating all ordered candidate pairs, the study examined whether relation classifiers developed with controlled hard-negative sampling could maintain effective discrimination under a naturally sparse exhaustive candidate distribution.

The findings revealed a substantial generalization gap between the sampled development setting and the sealed exhaustive evaluation. Both the TF--IDF + Linear SVM and weighted RoBERTa recovered genuine relations, but their predictions were accompanied by extensive false-positive proliferation when applied across the complete candidate space. The results therefore show that correct concept endpoints alone are insufficient for precise relation extraction and that performance measured under sampled-negative development conditions may not characterize behavior under severe relational sparsity.

The complementary SmolLM2 evaluation exposed additional challenges associated with end-to-end extraction, including missed relations, unsupported relations, and malformed structured outputs. Together, the discriminative and generative experiments demonstrate that errors can arise at multiple stages of structured knowledge extraction, from discovering and representing relational structures to distinguishing genuine relations among plausible concept pairs.

These findings position candidate-space management and false-positive control as central methodological challenges for relation extraction from aviation safety narratives. Future work should investigate larger relation-annotated resources, independently trained candidate-proposal mechanisms, evidence-grounded verification, and constrained or joint extraction architectures. Such approaches should be evaluated using strictly separated development and test resources and under candidate distributions that preserve the relational sparsity encountered when complete narratives are processed. This evaluation framework provides a controlled basis for developing more precise and robustly evaluated methods for structured knowledge discovery from aviation safety narratives.

\section*{Abbreviations}

\noindent
The following abbreviations are used in this manuscript:

\begin{table}[H]
\centering
\begin{tabularx}{\textwidth}{lX}
\toprule
\textbf{Abbreviation} & \textbf{Definition} \\
\midrule
AI & Artificial Intelligence \\
BERT & Bidirectional Encoder Representations from Transformers \\
CRF & Conditional Random Field \\
FP & False Positive \\
FN & False Negative \\
KG & Knowledge Graph \\
LLM & Large Language Model \\
MCC & Matthews Correlation Coefficient \\
NLP & Natural Language Processing \\
NTSB & National Transportation Safety Board \\
RAG & Retrieval-Augmented Generation \\
SVM & Support Vector Machine \\
TF--IDF & Term Frequency--Inverse Document Frequency \\
TN & True Negative \\
TP & True Positive \\
\bottomrule
\end{tabularx}
\end{table}


\end{document}